# Why is diagnosis using belief networks insensitive to imprecision in probabilities?

Max Henrion, Malcolm Pradhan[†], Brendan Del Favero[*], Kurt Huang[†], Gregory Provan, Paul O'Rorke[‡]

The Decision Laboratory
Institute for Decision Systems Research
4984 El Camino Real, Suite 110
Los Altos, CA 94022

## Abstract

Recent research has found that diagnostic performance with Bayesian belief networks is often surprisingly insensitive to imprecision in the numerical probabilities. For example, the authors have recently completed an extensive study in which they applied random noise to the numerical probabilities in a set of belief networks for medical diagnosis, subsets of the CPCS network, a subset of the QMR (Quick Medical Reference) focused on liver and bile diseases. The diagnostic performance in terms of the average probabilities assigned to the actual diseases showed small sensitivity even to large amounts of noise. In this paper, we summarize the findings of this study and discuss possible explanations of this low sensitivity. One reason is that the criterion for performance is average probability of the true hypotheses, rather than average error in probability, which is insensitive to symmetric noise distributions. But, we show that even asymmetric, logodds-normal noise has modest effects. A second reason is that the gold-standard posterior probabilities are often near zero or one, and are little disturbed by noise.

## 1 INTRODUCTION

Many researchers have criticized the use of probabilistic representations in general, and belief networks in particular, because of the need to provide many numbers to specify the conditional probability distributions. These numbers may be judged directly by experts in the domain of interest, or they may be obtained by fitting a belief network to a set of observed cases. In either case, they are liable to imprecision, due to the difficulty people have in expressing their beliefs numerically, or due to the finite amounts of case data available for estimation. If very precise numbers are needed, these criticisms would have considerable bite. If rough approximations are adequate, then they may lose their sting.

[†] also Section on Medical Informatics, Stanford University, Stanford, CA 94305.
[*] also Engineering-Economic Systems, Stanford University, Stanford, CA 94305.
[‡] The Cognition Institute, University of West Florida, Pensacola, FL 32514.

In recent research, we have explored empirically the question of how precise these probabilities need to be to obtain adequate performance from the belief network for diagnostic inference. Overall, we have been surprised at the degree of robustness of belief networks that we have found to imprecision in the probabilities. The primary goal of this paper is to explore possible reasons for this robustness so that we can better understand how it arises, when we can rely on it, and when we cannot. By better understanding how the diagnostic performance is affected by error or uncertainty in the probabilities, we can discover how precisely we need to estimate the probabilities, and hence whether or when belief networks can be practical and reliable.

## 2 FINDINGS OF ROBUSTNESS

Our investigation was based on a series of real-world Bayesian belief networks (BNs), rather than on the randomly generated, abstract knowledge bases (KBs) used in much of the experimental research to compare knowledge representations. Although it is easy to generate BNs with a wide range of different characteristics — such as ratio of arcs to nodes, ratio of source nodes to internal nodes, or frequency of directed cycles — we wanted to focus on BNs that have the characteristics of real application domains. We believe such BNs are more likely to be relevant to other real application domains than artificially-generated networks. The problem domain that we use in this study is medical diagnosis for hepatobiliary disorders (liver and bile diseases).

Numerical probabilities for belief networks may be estimated from empirical data or assessed by experts. In either case, the numbers are subject to various sources of inaccuracy and bias. For example, the data may be obtained from a sample that is not truly representative of the application domain, or the expert may have nonrepresentative experience. Limited sample sizes lead to random error. The process of expert assessment of probabilities is subject to a variety of inaccuracies which have been the subject of extensive study. [Kahneman, et al.1982] The question we wish to address here is how far these sources of imprecision are likely to matter.

We performed experiments to examine the sensitivity of BNs to the expert probabilities. In the experiments, we assessed the effects of random noise in the numerical probabilities on diagnostic performance, measured as the probability assigned to the correct diagnosis averaged over a large number of diagnostic test cases, for three different BNs. We added random noise to the probabilities derived from the standard, empirically derived mapping [Heckerman and Miller 1986] from frequency weights into probabilities. We added noise separately to the *link probabilities*, *leak probabilities*, and the *prior probabilities*. By examining the effect of noise separately on each of these three types of probability, we were able to differentiate among them in terms of their effect on diagnostic performance.

## 2.1 EXPERIMENTAL NETWORKS

We derived the experimental BNs from an early quasi-probabilistic KB, the Computer-based Patient Case Simulation (CPCS) [Parker and Miller 1987)]. CPCS uses a representation derived from the Internist-1 [Miller, et al. 1982] and Quick Medical Reference (QMR) [Miller, et al. 1986] expert systems. In these knowledge bases, causal links, such as relationships between diseases and findings are quantified as *frequency weights*, specifying the chance that one diseases will give rise to a finding or other variable, on a five-point qualitative scale. In previous work, our group developed a method to convert from the Internist-1 /QMR representation to a belief network representation, with specific independence assumptions — conditional independence of findings given diseases, noisy-OR influences of diseases on findings, and marginal independence of diseases [Shwe, et al. 1991]. Empirical comparison of QMR with the probabilistic reformulation, QMR-BN, demonstrated comparable diagnostic performance [Middleton, et al. 1991], even though some information (e.g. linkages between diseases) was not employed in QMR-BN.

Our first task in the current work was to convert the CPCS knowledge base into a coherent BN, mapping frequency weights into *link probabilities*, which are the conditional probabilities of each finding given each disease. We also had to assess additional *leak probabilities*, to quantify the chance that each finding, or other variable, will be present but not caused by one of the diseases or other variable in the knowledge base, and *prior probabilities* to quantify the prevalence rate of each disease or predisposing factor.

Because CPCS-BN is large and multiply-connected, it is impractical to perform inference with available inference algorithms using the entire network. If we wish to compute only the posterior probabilities of a small set of diseases, we can perform inference using only the subnetwork of the CPCS network that is relevant. We selected subnetworks from the full CPCS-BN using the BN graphical tool Netview [Pradhan, et al,1994]. Netview allows the user to display selected subsets of nodes from a network for simplicity of visualization and editing.

We extracted three subnetworks from the full CPCS-BN for the experiments, named BN2, BN3, and BN4, containing two, three, and four diseases. While CPCS has 365 nodes, BN2, BN3, and BN4 contain 42, 146, and 245 nodes, respectively.

## 2.2 TEST CASES

We needed far more test cases to estimate reliably the effects of the experimental manipulations on the diagnostic performance than the small number of cases available from real patient data. Accordingly, we generated sample test cases directly from the BNs themselves, generating findings according to the probabilities specified by the network using logic sampling [Henrion 1988]. We used the full CPCS network and the standard probability mapping for generating the test cases.

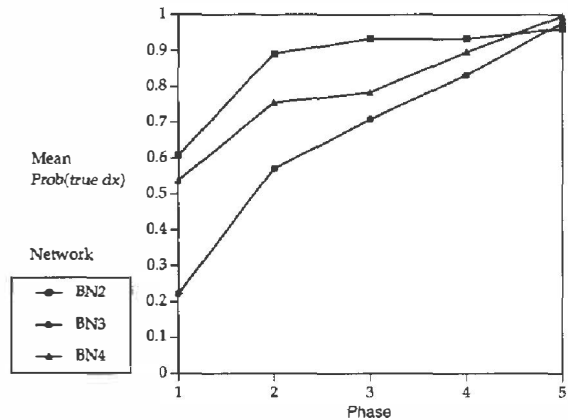

**Figure 1.** The average probability of the true diagnosis as a function of the Phase (amount of evidence) for each network without noise.

Since we wanted to investigate how the amount of evidence affects sensitivity to the experimental manipulations, we generated cases with varying numbers of findings. The test cases, as initially generated, include values for all findings. To create harder cases with fewer findings, and also for greater medical realism, we created five cases from each initial case, by revealing the findings in five Phases, approximating the order in which findings would be revealed in a real medical consultation. The five Phases correspond to successive stages in medical diagnosis. Phase 1 includes symptoms and findings volunteered by the patient — e.g., abdominal pain in the epigastrium. Phase 5 includes all previous phases plus expensive, invasive laboratory tests, including pathology findings, which are usually obtained through biopsies — e.g., hepatocellular inflammation and/or necrosis.

## 2.3 MEASURES OF DIAGNOSTIC PERFORMANCE

We quantify diagnostic performance as the probability assigned by each network to each true diagnosis, averaged over the set of test cases. We analyze separately the probabilities assigned to each disease when present (we call this the true positive rate), and the probability assigned to the absence of each disease when absent (we call this the true negative rate).

Figure 1 plots the average probability assigned to the true diagnosis (true positive and true negative) as a function of the Phase for each of the three networks. As expected, diagnostic performance improves consistently with Phase — that is, the additional findings available in the later Phases lead to a higher average posterior probability of the true diagnosis. Performance starts out relatively poorly for Phase 1, especially for BN3, where the average posterior probability for the true diagnosis is 0.22. But, with with the entire set of evidence available in Phase 5, diagnostic performance becomes excellent, averaging 0.978 over the three networks.

## 2.4 LOG-ODDS NORMAL NOISE

Perhaps the most obvious way to add noise to a probability is to add a random noise directly to the probability. This approach has two problems. First, a large additive error is likely to produce a probability greater than 1 or less than 0, and so needs to be truncated. Second, an error of plus or minus 0.1 seems a lot more serious in a probability of 0.1, ranging from 0 to 0.2, than it does in a probability of 0.5, ranging from 0.4 to 0.6. Link probabilities near 0 or 1 can have enormous effects in diagnosis for findings that are present or absent (respectively).

A more appealing approach that avoids these problems is to add noise to the log-odds rather than the probability. This approach can be viewed as a version of Fechner's law of psychophysics in which similar just-noticeable differences in quantities such as weight or brightness are approximately constant when measured on a logarithmic scale. Since probability has two bounds, 0 and 1, we wish to have a symmetric effect near each bound. The log-odds transformation provides exactly this behavior.

More specifically, we transformed each probability $p$ into log-odds form, added normal noise with a standard deviation of $\sigma$ and transform back into probabilities. We define the log-odds transformation as:

$$Lo(p) = \log_{10}[\, p/(1-p)\,] \qquad (1)$$

We add log-odds noise to the probability as follows:

$$p' = Lo^{-1}[\, Lo(p) + e\,], \text{ where } e = Normal(0,\sigma) \qquad (2)$$

Note that the amount of noise implied by Figure 1 is very substantial. Figure 2 shows second-order probability density functions for logodds-normal probability distributions for sigma 0.1, 0.2, 0.5, 1, 2, and 3. For $\sigma \le 0.5$, the density function is unimodal. For $\sigma \ge 1$, the density function becomes bimodal with peaks at zero and one. In other words, for $\sigma$ values of 1 or more, the noise is so extreme that most of the probabilities are near zero or one, rather than near the gold-standard probability. The logodds-normal distribution guarantees only that the median probability is equal to the gold-standard. The mean probability may be very different.

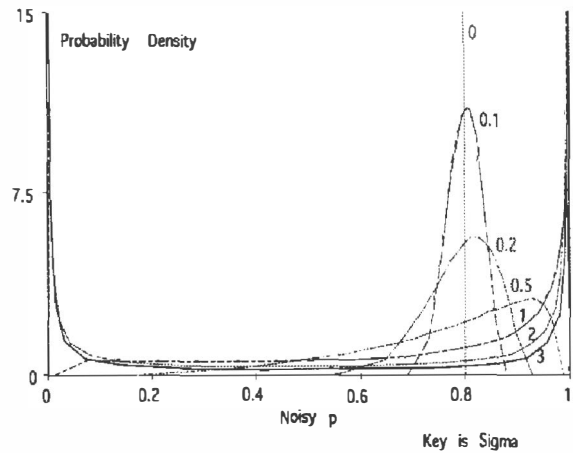

**Figure 2.** Effect of adding Log Odds Normal noise to a probability of 0.8.

## 2.5 EXPERIMENTAL DESIGN

We start with binary networks using the standard mapping with no noise ($\sigma = 0$), and then add noise, generated independently for each link probability in the network, with $\sigma = 1.0$, $\sigma = 2.0$, and $\sigma = 3.0$. We generated 10 noisy networks independently for each $\sigma$. Similarly, we created noisy networks adding noise only to the leak probabilities, and only to the prior probabilities for each network.

The total number of networks used in this experiment were 273, comprised of 3 levels of noise x 3 probability types (link, leak, and priors) x 10 samples x 3 networks, plus the original 3 standard networks without noise. For each of these networks, we ran the entire set of cases, requiring a total of 291,200 runs. We assessed performance using the average probability assigned to the true diagnoses.

## 2.6 RESULTS

Figure 3 plots the average performance — the probability assigned to the true diagnosis — for the four-disease network against the four levels of noise on the link, leak, and prior probabilities. Plots for the two-disease and three-disease networks are similar. We see that, as expected, increasing noise consistently degrades performance for each type of probability — link, leak, and prior. Performance is relatively more sensitive to noise on links than to noise on priors or leaks. The effect of noise on the leaks and priors is indistinguishable for networks BN3 and BN4.

The introduction of noise in the numerical probabilities does degrade performance, as expected. However, the amount of degradation is surprisingly small when one considers the degree of noise. It appears that even large errors in the probabilities produces only modest degradations in performance

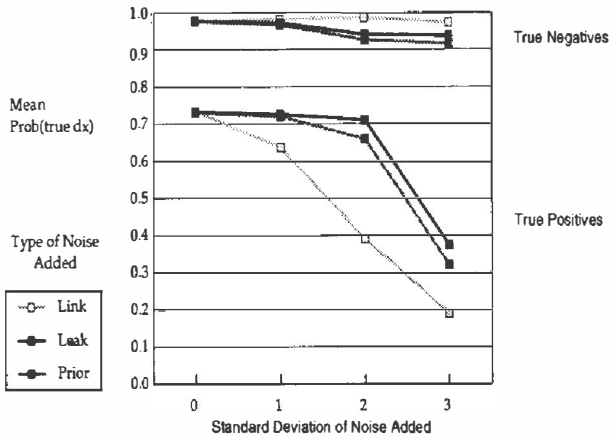

**Figure 3.** Effects of noise in the link, leak, and prior probabilities on average diagnostic performance for the BN4 network.

### 2.7 EFFECTS OF NOISE ON TRUE POSITIVES AND NEGATIVES

Hitherto, our analysis has combined the probabilities assigned to true positives (TP) — i.e., the probability of the disease for cases in which the disease is present, and probabilities assigned to true negatives (TN) — i.e., the probability of no disease for cases in which the disease is absent. We can obtain interesting insights that help explain our results by examining the effects of noise on these two measures separately. Figure 3 plots the average probability assigned to the true diagnosis separately for TP and TN, as a function of the noise level in the link, leak, and prior probabilities. These results were similar for each of the three networks. Accordingly, for simplicity, Figure 3 shows results averaged over the three networks.

The first point to note is that, without noise, the average performance for true negatives (TN) at 0.97 is substantially better than for true positives (TP) at 0.73. In other words, the system is more likely to miss a disease that is present than to falsely diagnose a disease that is not present. This tendency to underdiagnose should be expected because the prevalence of diseases in the test cases is much larger than would be expected according to the prior probabilities on the diseases. Note that we deliberately generated most of the test cases to contain one or more diseases to provide more information on diagnostic performance on interesting cases, even though according to the priors, more cases would have no diseases.

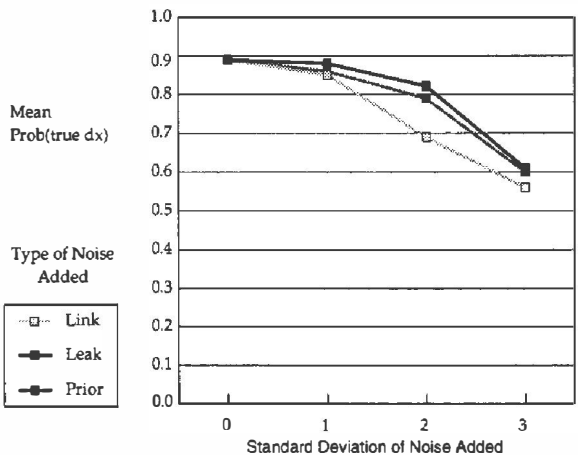

**Figure 4.** Effect of noise in the link, leak, and prior probabilities on the true positive and negative probabilities, averaged over BN2, BN3, and BN4.

Now let us look at the effect of noise levels on TP and TN. Noise in the link probabilities significantly degrades performance for TP, but has no statistically detectable effect on TN ($\alpha = 0.05$). Conversely, noise in the leak probabilities has no statistically detectable effect on TP at noise levels $\sigma = 1$ and $\sigma = 2$, but leak noise significantly degrades TN. Finally, noise in the priors has a similar, slight, but significant, effect in degrading performance on both TP and TN. At the highest noise setting, $\sigma = 3$, the performance of networks with leak noise and prior noise sharply decline because the disruption to the probability values is so extreme (Figure 2).

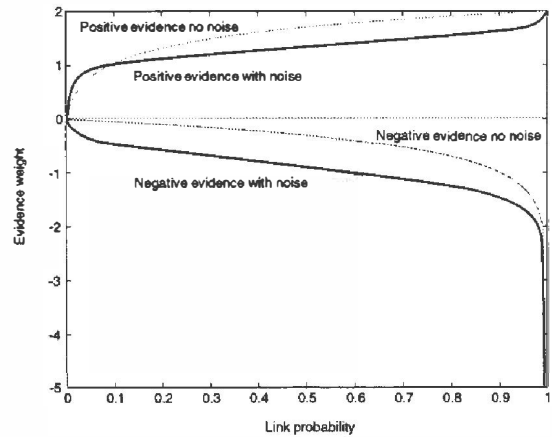

**Figure 5.** Evidence weights for a disease from a single positive and negative finding as a function of the link probability, without noise and with sigma=2 noise. Note that noise tends to decrease the evidence weight for both positive and negative findings.

Why should link noise and leak noise show these contrary effects on TP and TN? We can explain these results by analyzing the role of the link and leak probabilities in the diagnosis. For simplicity, let us consider the effect of a single finding $F$, being present, $f$, or absent, $-f$, on the posterior odds of a single disease $D$. A standard measure of the strength of diagnostic evidence is the log-likelihood ratio, also known as the evidence weight:

$$EW(f, D) = \log_{10}\left[\frac{P(f|d)}{P(f|\neg d)}\right]$$

$$EW(\neg f, D) = \log_{10}\left[\frac{P(\neg f|d)}{P(\neg f|\neg d)}\right]$$

$P(f|d)$, the probability of the finding when the disease is present is:

$$P(f|d) = 1 - (1 - Leak(F))(1 - Link(D, F))$$
$$= Leak(F) + Link(D, F)(1 - Leak(F))$$

$P(f\backslash\neg d)$, the probability of the finding when the disease is absent, is the leak probability, $Leak(F)$. We can now rewrite the likelihoods in terms of the link and leak probabilities.

$$EW(f, D) = \log_{10}\left[\frac{leak(F) + link(D, F)(1 - leak(F))}{leak(F)}\right] \quad (3)$$

$$EW(\neg f, D) = \log_{10}[1 - link(D, F)] \quad (4)$$

Notice the leak probability does not play a role in the negative evidence weight (Eq. 4), because if a finding is absent then the leak must be 'off' by definition.

Figure 5 plots the evidence weights for positive and negative findings, as a function of the *link* probability, and the mean evidence weight with 2 sigma noise in the link probability. It demonstrates that, on the average, noise in the link decreases the evidence weight for the finding. This effect arises from the fact that the evidence weights are concave functions of the link probability. Accordingly, the noise in the links will tend to reduce the probability assigned to the true positive, reducing performance as noise increases. Noise in the links, by reducing the evidential strength of findings can only increase the probability assigned to true negative, but this effect is undetectable because the true negative rate is already high.

The impact of noise on the positive evidence (Eq. 3) is bounded by the value of the leak. The smaller the leak, the greater the possible effect on the positive evidence. In contrast, the negative evidence weight (Eq. 4) can be significantly decreased if the link probability is close to 1.0, as is the case with sensitive findings.

A related argument demonstrates that noise in leak probabilities will tend to increase the strength of evidence on the average. In consequence, noise in leaks also tends to increase false positives and so degrades performance for true negatives. The effect on true positives is again not detectable.

## 3 WHY SYMMETRIC UNCERTAINTY IN PROBABILITY DOES NOT MATTER

One reason that we may find these findings of robustness surprising may be due to a misapprehension of the nature of uncertainty about probability. Random noise in probability tends to be much less important than random noise in other quantities of interest, because, when averaged over many cases, we should care only about the average or expected probability. This is reflected in the scoring rule that we used in our study, the average probability assigned to the disease that is present, or to its absence if the disease is absent. We should not care about imprecisions in estimating the probabilities or other sources of noise in the system if they do not affect the expected probability. In other words, any symmetrical distribution of noise around the "gold standard" probability will not affect the score of the system.

For example, suppose that in a set of cases with findings $F$, the probability of disease $D$ being present, $P(D|F) = 0.7$, and that a diagnostic expert system A, computes a posterior probability for $D$ of $P_A(D|F)$. The score for this performance, computed as above, as the probability assigned to the correct diagnosis, is given by:

$$S(A) = P(D)|F) P_A(D|F) + [1 - P(D|F)][1 - PA(D|F)] \quad (5)$$

If we assume that $A$ is well-calibrated, so that $P_A(D|F) = 0.7$, the score is

$$S(A) = 0.7 \times 0.7 + (1 - 0.7)(1 - 0.7)$$
$$= 0.49 + 0.09 = 0.56 \quad (6)$$

Now consider a diagnostic expert system, $B$, which is subject to severe random noise, such that it estimates the posterior probability of $D$ half the time $P_B(D|F)=1$ and half the time $P_B(D|F)=0.4$. Note that $B$ is still well-calibrated because the expected probability given by $B$ over the noise is given by $E[P_B(D|F)] = 0.5 \times 1 + 0.5 \times 0.4 = 0.7$. The score for $B$ will be:

$$S(B) = E[P(D|F) \times P_B(D|F) + [1 - P(D|F)][1 - P_B(D|F)]] \quad (7)$$

$$S(B) = P(D|F) \times E[P_B(D|F)] + [1 - P(D|F)][1 - E[P_B(D|F)]] \quad (8)$$

$$S(B) = 0.7 \times 0.7 + [1 - 0.7][1 - 0.7] = 0.56 \quad (9)$$

In other words, the score is the same for expert systems $A$ and $B$. More generally, $A$ and $B$ will have the same score provided the noise to which $B$ is subject does not change the expected value of the posterior probability estimated by $B$, i.e.

$$E[P_B(D|F)] = P_A(D|F) \quad (10)$$

Therefore, any kind of noise in $B$'s estimates $P_B(D \mid F)$ that does not affect the expectation of the estimate will not degrade the performance of $B$ using an average measure. In some cases, of course, $B$ may do worse than $A$ — but in other cases it will do better. On the average it will do the same. In this way, probability is unlike other quantities of interest — such as, a patient's blood pressure or the net present value of an investment — where imprecision in the measurement may affect the utility of the measurement even if it does not affect the expected value.

## 4 EFFECT OF LOGODDS NORMAL ERROR

If symmetric uncertainty about probabilities has no effect on results, what effect might asymmetric uncertainty have? There is reason to believe that uncertainty about probabilities may often be asymmetric. For example, an expert may believe that a finding, $F$, is very probable given a disease, $D$, but not certain. Suppose the expert provides $P(F|D)=p$, with $p=0.9$ as her best estimate of the conditional probability, but feels quite uncertain. Her belief that the relationship is not certain implies that $p < 1.0$. However, she may well feel that $p=0.7$ would not be surprising. The tight upper bound and loose lower value suggest an asymmetric distribution.

We can also provide a mathematical argument for asymmetric posterior probabilities for diagnosis with many uncertain sources of evidence. Suppose that we have $n$ findings $F_i$ for $i=1$ to $n$ that are conditionally independent, given a disease $D$. Using the standard log-odds formulation of Bayes' rule, we have:

$$Lo(D \mid F_1, F_2 ... F_n) = \qquad (11)$$

$$Lo(D) + \sum_{i=1}^{n} \left\{ Log_{10}[P(F_i \mid D)] - Log_{10}[P(F_i \mid \overline{D})] \right\}$$

Let us view the uncertain link probabilities $P(F_i \mid D)$ and $P(F_i \setminus D)$ as random variables. The posterior logodds is therefore the sum of differences between $n$ pairs of random variables. By the central limit theorem, the distribution over the logodds posterior will tend with increasing $n$ to a normal distribution. Equivalently, the posterior odds will tend to a lognormal distribution and the posterior probability will tend to a logodds-normal distribution. This is one of the reasons we used logodds-normal noise in the experiments described above. The logodds-normal distribution is, in general, asymmetric in probabilitty.

Let us examine the effect of a logodds-normal uncertainty in posterior probabilities on the diagnostic performance of a belief network. Figure 6 shows the expected error, that is the absolute average difference in scores (probability assigned to true diagnoses) between "gold standard" probability and the noisy probability, as a function of the gold standard. The noise is applied as a logodds-normal with σ varying beween 0 and 3, as shown on the key.

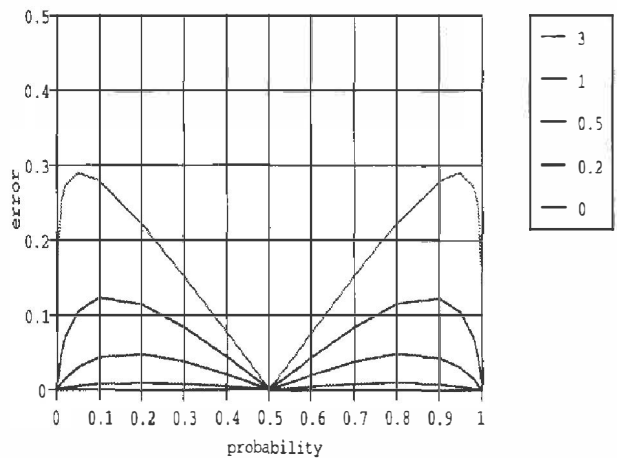

**Figure 6.** Effect of Log Odds Normal noise on diagnostic performance.

For a gold-standard probability of 0.5, the expected error is zero. This result arises because the logodds-normal distribution is symmetric around 0.5. The expected error is also zero for gold standard probability near 0 or 1. The expected error is largest for intermediate probabilities. The largest expected error for σ = 0.3 is about 0.021 at about $p=0.25$ and $p=0.75$. Larger expected errors occur for at extreme noise levels of σ = 1.0 and σ = 3.0. As we have mentioned before, noise levels of σ ≥ 1.0 are very extreme. For such values of σ, the second-order probability density functions are bimodal, with peaks at 0 and 1 Figure 2

## 5 CONCLUSIONS

A better understanding of sensitivity to errors or noise in numerical probability can help guide the builder of belief networks in deciding how much effort it is worth putting into probability assessment—whether probabilities are assessed directly by experts, or estimated empirically from collected patient case data. It could also help us understand what levels of precision in diagnosis we can expect given the inevitable imprecision in the input probabilities. A better understanding of the relative sensitivity to links, leaks, and priors could help guide the knowledge engineer in allocating effort in assessing these three classes of probability.

In this paper, we have examined the sensitivity of several belief networks on diagnostic performance to imprecision in the representation of the numerical probabilities. Overall, we have found a surprising level of robustness to imprecision in the probabilities. Here we summarize the key findings, explore their implications, and discuss their limitations.

The addition of massive amounts of random noise to the link, leak, and prior probabilities produced only modest decrements in diagnostic performance. Noise in the link probabilities had the largest effect in reducing performance for all three networks. Noise in the leak and prior probabilities had smaller effects, but performance consistently degraded with the level of noise for all three networks.

The surprisingly small effect of large amounts of random noise should be reassuring for those constructing belief networks. The results presented here—together with additional results from studies [Pradhan, Henrion, Provan, Del Favero, and Huang, 1996] of different mappings, reductions in the number of levels used (e.g., severities of diseases and findings), and studies of the impact of unrepresented nodes (e.g., diseases not present in the CPCS subnetworks)—provide empirical evidence that it is much more important to obtain the correct qualitative information. It is more important to identify findings, and diseases, IPSs, and their relationships, than to quantify the relations with a high level of precision. Experience suggests that domain experts are much more comfortable providing these kinds of qualitative knowledge than they are providing quantitative probabilities, although use of probability elicitation methods can make the latter more acceptable. Knowledge that high levels of precision are not necessary should greatly improve acceptance of these techniques.

Ultimately the purpose of any diagnostic system is to lead to better decisions—more cost-effective treatments of diseases, or repair to complex artifacts. In this paper, we have measured performance by accuracy of diagnosis, not by improved decisions. However, if imprecision in the representation does not degrade the diagnosis, it should not degrade the decision. In general, diagnostic accuracy is more sensitive to imprecision in the model of system being diagnosed than is the quality of the decision. Therefore, where we find that the quality of diagnosis is robust to imprecision, we can be confident that the quality of decisions will be equally or more robust.

While we believe that these results provide intriguing and suggestive evidence, we should caution that they should not be viewed as definitive for all BNs. First, note that these results are for a diagnostic application. There is reason to believe that predictive applications may show greater sensitivity. Second, these networks, like most existing large BNs, use noisy-OR influences, or their generalization to noisy MAX influences. In fact, in most diagnostic belief networks constructed hitherto, the large majority of influences *are* noisy OR links. But, BNs that make extensive use of other types of influence may show different sensitivities.

In addition, our results are specific to the types of noise and the measure of diagnostic performance studied here. It would be interesting to study noise intended to mode systematic cognitive biases that have been observed in psychological studies of probability elicitation and human reasoning under uncertainty.

It may be possible to improve on our measure of performance. Our experiments and our performance measure have a major advantage over some alternatives that gauge consistency with a diagnostic network's assessment of probability. Because we generate the test data, we know whether diseases are actually present in the test cases. Our measure reflects "real" diagnostic accuracy in that it takes account of the true diagnoses rather than relying completely on subjective assessments, such as might be provided by an expert. On the other hand, our measure is not a proper scoring rule. (A scoring rule is proper if it maximizes the expected score of an assessor when they report their true assessment of probabilities.) Since our measure is a linear score, it can reward assignments by noisy networks that disagree with the assignments of noise-free networks. Also, unlike some scoring rules, (e.g., the Brier score), our measure does not separate different aspects of performance such as calibration errors and discrimination between different levels of probability. It may also be worthwhile to develop methods to account for differences between the frequencies that are present in the test data and the empirical frequencies used to construct the network and the frequencies that are seen in actual diagnostic practice.

Clearly, there is a need for additional work to explore these possibilities. While we believe that further experimental work is essential, we expect that theoretical analysis will also help to provide a deeper understanding of some of the findings, and suggest profitable avenues for further experimentation.

We are not the first to argue that the conclusions of diagnostic and other expert systems may have low sensitivity to imprecision in the numerical parameters. However, in heuristic representations where both the structural assumptions embody unexplicated simplifications of principles of rationality, it is often hard to separate the question of numerical approximations from structural simplifications. In the context of a probabilistic belief network, it is possible to be clear about both structural simplifications, such as independence assumptions, and the effects of numerical approximation, and so differentiate among these potential sources of error, in a way that is impossible in heuristic representations of uncertainty.

Our results lend support for the value of qualitative probabilistic representations, such as the QPNs [Wellman 1990, Henrion and Druzdzel 1990] and infinitesimal probability schemes [Goldszmidt and Pearl 1992]. Indeed, we have performed some initial experimental comparisons of the performance of a BN for machine diagnosis using a qualitative infinitesimal representation (the $\kappa$ calculus) with a numerical BN. We found little difference in diagnostic performance between the numerical and infinitesimal representations for cases with small fault priors [Henrion, et al. 1994]. The findings we have presented here help to explain the small differences between the qualitative and quantitative representations.

# 6 ACKNOWLEDGEMENTS

This work was supported by NSF grant IRI 91-20330 to the Institute for Decision Systems Research. We would like to thank Joseph Kahn for providing feedback and commentary on earlier drafts, Dr. Blackford Middleton for his help in developing CPCS-BN, and Lyn Dupré for her editorial help.